\documentclass{article}

\usepackage{microtype}
\usepackage{graphicx}
\usepackage{subfigure}
\usepackage{ulem}
\usepackage{booktabs} 

\usepackage{hyperref}
\usepackage{hhline}
\usepackage{multirow}

\usepackage[accepted]{style}
\usepackage{amsmath} 
\usepackage{amssymb}
\usepackage{cleveref}
\usepackage{threeparttable}

\newcommand{\vz}[1]{\mathsf{z}_{#1}}
\newcommand{\vr}[1]{\mathbf{r}_{#1}}

\newcommand{\RR}{\mathbb{R}}
\newcommand{\ZZ}{\mathbb{Z}}
\newcommand{\NN}{\mathbb{N}}

\newcommand{\mlp}{\mathsf{NN}}

\def\DFT{\ensuremath{\mathsf{DFT}}}
\def\star{^\ast}

\DeclareMathOperator*{\argmin}{argmin}

\icmltitlerunning{Reducing the Cost of Quantum Data By Backpropagating Through Density Funcitonal Theory}

\begin{document}

\twocolumn[


\icmltitle{Reducing the Cost of Quantum Chemical Data By \\Backpropagating Through Density Functional Theory}



\icmlsetsymbol{equal}{*}

\begin{icmlauthorlist}
\icmlauthor{Alexander Mathiasen}{gc}
\icmlauthor{Hatem Helal}{gc}
\icmlauthor{Paul Balanca}{gc}
\icmlauthor{Adam Krzywaniak}{gc}
\icmlauthor{Ali Parviz}{mila}
\icmlauthor{Frederik Hvilsh\o j}{}
\icmlauthor{Blazej Banaszewski}{gc}
\icmlauthor{Carlo Luschi}{gc}
\icmlauthor{Andrew William Fitzgibbon}{gc}
\end{icmlauthorlist}

\icmlaffiliation{gc}{Graphcore}
\icmlaffiliation{mila}{Mila - Québec AI Institute}

\icmlcorrespondingauthor{Alexander Mathiasen}{alexander.mathiasen@gmail.com}

\icmlkeywords{Machine Learning, ICML}

\vskip 0.3in
]

\definecolor{darkgreen}{RGB}{0,150,0}
\definecolor{darkred}{RGB}{150,0,0}


\printAffiliationsAndNotice{}  

\begin{abstract}
Density Functional Theory (DFT) accurately predicts the quantum chemical properties of molecules, but scales as $O(N_\text{electrons}^3)$. 
\citet{schnorb} successfully approximate DFT 1000x faster with Neural Networks (NN). 
Arguably, the biggest problem one faces when scaling to larger molecules is the cost of DFT labels. 
For example, it took years to create the PCQ dataset \cite{pcq} on which subsequent NNs are trained within a week. 
DFT labels molecules by minimizing energy $E(\cdot)$ as a ``loss function.''
We bypass dataset creation by directly training NNs with $E(\cdot)$ as a loss function. 
For comparison, \citet{schnorb} spent 626 hours creating a dataset on which they trained their NN for 160h, for a total of 786h; our method achieves comparable performance within 31h. 
\end{abstract}

\section{Introduction}
Density Functional Theory (DFT) accurately predicts the quantum mechanical properties of molecules
given their atom types and atom positions, with applications in the design of drugs \cite{proteinligand} and materials \cite{opencatalyst}.
Accurate DFT can be enormously compute-intensive, so effort has been devoted to train Neural Networks (NNs) to efficiently approximate DFT.
However, the high computational cost of DFT means that it is prohibitive to generate enough data to unlock the full benefit of NN scaling laws ~\cite{scalinglaws_language,scalinglaws_materials}. 
The largest existing datasets contain no more than $10^9$ examples, computed using a low resolution DFT (a minimal basis set) \cite{qm1b}. 
These dataset sizes are almost certainly insufficient to scale molecular NNs to the magnitude of Large Language Models (LLMs),  with current models no larger than a billion parameters, often less than 10M \cite{phisnet,qhnet}.

\begin{figure}[th]
    \centering
    \includegraphics[width=0.48\textwidth]{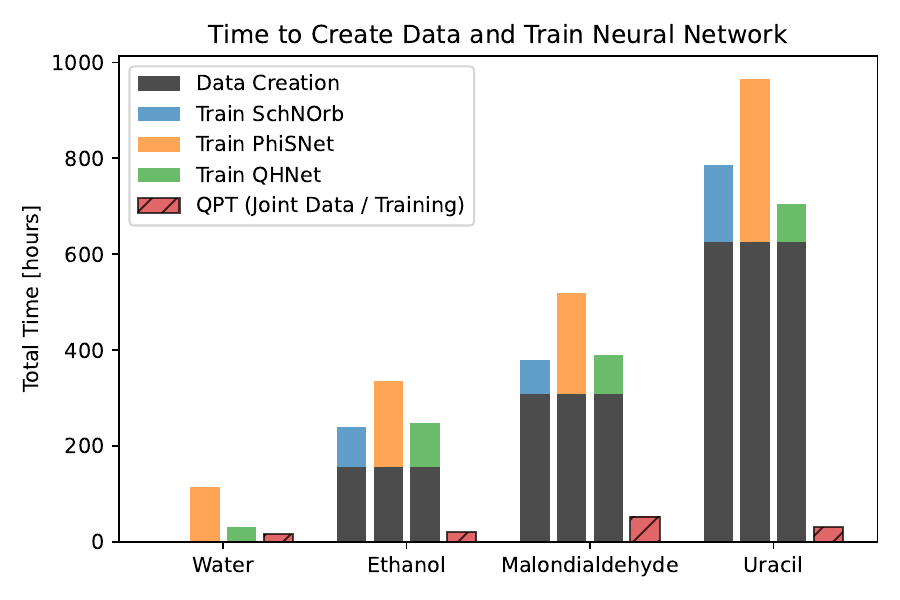}
    \vspace{-7mm}
    \caption{The time to label molecules with Density Functional Theory (DFT) scales as $O(N_{\text{electrons}}^3)$. 
    The cubic scaling makes it impractical to create datasets with molecules like peptides or proteins with larger $N_{\text{electrons}}$. 
    Our Quantum Pre-trained Transformer (QPT) bypasses the expensive DFT labeling by training with ``DFT's loss function'', the energy $E(\cdot)$. 
    This circumvents the cost of creating datasets, paving a way to scale both the size of molecules~and~NNs. 
    }
    \label{fig:total_time}
    \vspace{-2mm}
\end{figure}

We bypass the expensive DFT labeling by introducing a new pre-training technique. 
DFT labels molecules by minimizing the energy $E(\cdot)$ as a ``loss function,'' which require up to 50 DFT iterations to converge. 
Instead, we train our NN with $E(\cdot)$ as a loss function, which can be computed with 1~DFT iteration.  
Our Quantum Pre-trained Transformer (QPT) obtains comparable accuracy to prior work \cite{schnorb,phisnet,qhnet}
on the Hamiltonian version of MD17, while eliminating the dataset creation cost (see \Cref{fig:total_time}). 
Notably, the pre-training task is universal, in the sense that it allows one to compute any property that could be computed from DFT (e.g., forces, molecular orbital energies, homo-lumo gap, etc.). 
During training, it takes us 0.41s to compute and backpropagate through $E(\cdot)$, more than $200\times$ faster than the 88s \citet{schnorb} spent creating a training example (see \Cref{fig:traintime}). 
This allows us to evaluate $E(X_i)$ on a new molecule $X_i$ each step of training, generating an infinite stream of data,
thereby mitigating overfitting and paving a way to arbitrarily scale model size.
\begin{table*}[h!]
\centering
\begin{tabular}{|l|l|l|l|l|l|l|l|l|}
\hline 
& & \multicolumn{3}{c|}{\textbf{Train}} & \multicolumn{4}{c|}{\textbf{Data Creation $+$ Train}} \\
                   \textbf{Molecule} & \textbf{Data Creation}                      & \textbf{SchNOrb} & \textbf{PhisHNet} & \textbf{QHNet} & \textbf{SchNOrb} & \textbf{PhisHNet} & \textbf{QHNet} & \textbf{QPT} \\
\hline
Water           & -    & -  & 112h & 30h & -    & -    & -    & \textbf{16.3h} \\
Ethanol         & 157h & 82h  & 179h & 90h & 239h & 356h & 247h  & \textbf{20.1h} \\ 
Malondialdehyde & 309h & 71h  & 210h & 80h & 380h & 519h & 389h  & \textbf{52.0h} \\
Uracil          & 626h & 160h & 339h & 79h & 786h & 965h & 705h  & \textbf{30.2h}\\
\hline
\end{tabular}
\caption{Prior work \cite{schnorb,qhnet} spends more time creating quantum data than training neural networks, e.g., \citet{schnorb} spent 626h creating the Uracil data but only trained for 160h. 
We drastically reduce the joint time to create data and train neural networks, while achieving comparable accuracy (see \Cref{table:accuracy}). 
We conjecture our approach will allow future work to arbitrarily scale the parameters of neural approximations to DFT. }
\label{table:time}
\end{table*}

\subsection{Background and notation}
\def\phistar{\phi\star}

The core of DFT is to compute the energy-minimizing electron configuration given a set of atoms $X = \{x_m\}_{m=1}^M$, where the $m^\text{th}$ atom $x_m = (\vz m,\vr m)$ is represented by its atomic number $\vz m\in\NN$ and atom position $\vr m\in\RR^{3}$.
The electron configuration or  equivalently the density function, 
can be parameterized by the Hamiltonian 
$H$,
in terms of which we define
\begin{equation}\label{equ:dft}
\DFT(X) := \min_H E(X; H), 
\end{equation}
where $E(X; H)$ is the energy of molecule~$X$ with electron configuration (density function) parameterized by~$H$. 
It is common to represent $H\in\RR^{N_{AO}\times N_{AO}}$ where $N_{AO}$ is the number of (user specified) atomic orbitals. Following common practice, we use Gaussian Type Orbitals (GTOs) \cite{pople1999nobel}. 

For many applications, the energy {\it per se} is not the quantity of interest, but the Hamiltonian itself
\begin{equation}
H^*(X):=\argmin_H E(X; H),
\end{equation}
from which useful molecular properties can be derived. 
This motivated prior work \cite{schnorb,phisnet,qhnet} to train $\mlp(X;\theta)$ to approximate $H^*(X)$: 
\begin{equation}
\mlp(X;\theta) \approx H^*(X)
\end{equation}
\def\ntrain{D}
Given a dataset of $D$ molecules $\{X_i\}_{i=1}^{\ntrain}$, training consists of minimizing over the neural network parameters~$\theta$:
\def\loss#1#2{\bigl| #1 - #2 \bigr|}
\begin{align}
\theta\star &= \argmin_\theta \sum_{i=1}^{\ntrain} \loss{\mlp(X_i; \theta)}{H^*(X_i)}_1
\label{eqn:classicml}
\end{align}
where the labels $H^*(X_i)$ are precomputed with DFT \cite{nabladft,qh9}.\footnote{The loss, while written as a 1-norm, may be any loss function, e.g., \citet{qhnet} and \citet{phisnet}  use the sum of Mean Absolute Error (MAE) and Root Mean Squared Error (RMSE). } 

In this paper, we are concerned with reducing the total time taken to train a new model, including dataset generation.
To see how we might find such a reduction, we 
insert the definition of $H^*(X)$ into the loss:
\begin{align}
\theta\star = \argmin_\theta \sum_{i=1}^{\ntrain} \loss{\mlp(X_i; \theta)}{
\underbrace{\argmin_H E(X_i, H)}_{\textcolor{gray}{\text{Takes 50 DFT iterations}}}
},
\end{align}

Notably, each of the $D$ optimization problems with respect to $H$ can be solved in parallel:
\begin{equation}
    H_1^*,..., H_D^*=\argmin_{H_1, ...,H_D} \sum_{i=1}^D E(X_i; H_i).
    \label{equ:hstars}
\end{equation}
Instead of learning from the solutions to \Cref{equ:hstars}, we propose to learn directly from $E(\cdot)$, i.e., 
while existing losses~(\ref{eqn:classicml}) encourage $\mlp(X,\theta)$ to ``look like''~$H$, we define a loss which encourages $\mlp(X,\theta)$ to ``act like"~$H$, which we call the {\it implicit DFT loss}: 
\begin{equation}
\theta^*= \argmin_\theta \sum_{i=1}^{\ntrain} \underbrace{E(X_i, \mlp(X_i;\theta))}_{\textcolor{gray}{\text{Takes 1 DFT iteration}}},
\label{equ:implicitdft}
\end{equation}

which can be trained using conventional machine-learning optimizers, but at considerably lower cost, as the optimizations over~$H$ and~$\theta$ are interleaved.
By analogy with the performance of stochastic gradient descent, where weight updates are made over many small batches per epoch, this implicit DFT loss allows an optimal~$\theta$ to be found with significantly less computation.

One may view the ``implicit DFT loss'' as unsupervised learning, in the sense that there are no labels $y_i$. 
That said, we remark that learning to approximate DFT is different to the normal machine learning framework. 
The goal of supervised learning \cite{learningfromdata} is to approximate an unknown target function $f:X\rightarrow Y$ given examples of the function $(x_i, f(x_i))$. With DFT, the function $f=DFT$ we seek to approximate is known, differentiable, and can be implemented  efficiently on hardware accelerators. In this paper, we advocate using these properties to move beyond supervised learning. 
\clearpage 

\paragraph{Why train NN to approximate DFT?} Why not just use the existing DFT libraries? 
Previous works explore different advantages.
The simplest advantage is that NN can accurately approximate DFT 1000x faster \cite{schnorb,phisnet,qhnet}, often attributed to the cubic $O(N_{AO}^3)$ scaling of DFT. 
Another answer, is that performance on small biological datasets appear to improve when training on large amounts of quantum data \cite{graphium}. 
One might imagine a variant of AlphaFold \cite{alphafold} with 10B parameters (instead of just 100M), pre-trained to predict the quantum chemical structure of proteins-ligand conformations, then subsequently fine-tuned on the 200k protein databank \cite{pdb}. 
Creating the dataset for such a pre-training run is prohibitive. 
We present a training technique which bypasses the cost of dataset creation.

\paragraph{Contributions.}
\begin{itemize}
    \item We bypass the expensive DFT labeling by training $\mlp$ with $E(\cdot)$ as a loss function. 
    \item We can generate a new training example each step of training. This paves a way to arbitrarily scale the size of molecular foundation models. 
    \item We demonstrate a Transformer applied to molecules with ``the fewest possible modifications,'' \cite{vit} achieve comparable performance to prior work by scaling to just 300M parameters. 
\end{itemize}

\section{Quantum Pre-trained Transformer}
Inspired by ViT \cite{vit} we apply a standard
Transformer directly to molecules ``with the fewest possible modifications.''

\begin{figure}[t]
    \centering
    \includegraphics[width=0.47\textwidth]{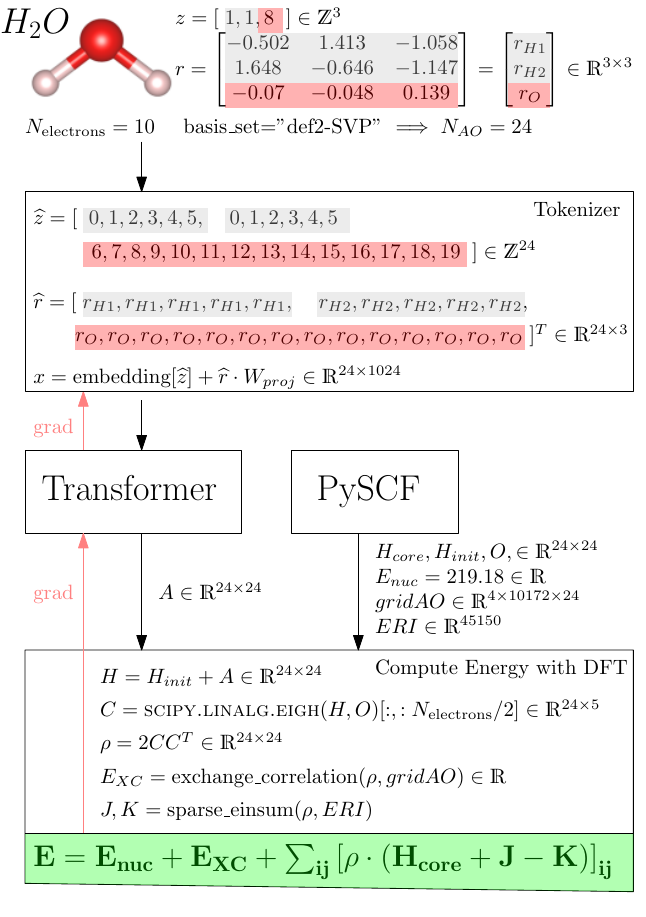}
    \vspace{-2mm}
    \caption{Visualization of how a molecule gets tokenized so our Transformer can process it and output an appropriately shaped matrix for the DFT energy computation.  }
    \label{fig:arch}
\end{figure}

We want the Transformer to use DFT as a loss function while dealing with differently sized molecules. 
DFTs energy function $E(\cdot)$ takes a square matrix of size $N_{AO}\times N_{AO}$ as input. 
Our Transformer outputs a $N_{AO}\times N_{AO}$ matrix by using the final pre-softmax attention head $A\in\mathbb{R}^{L\times L}$ where $L$ is the sequence length. 
We then tokenize the input molecule into atomic orbitals so $L=N_{AO}$. For~example, \citet{schnorb}, \citet{phisnet} and \citet{qhnet} use the DFT option (basis set) def2-SVP, for which hydrogen and oxygen are represented with 5 and 14 atomic orbitals respectively, so we would have 5 hydrogen tokens $H_1,...,H_5$ and 14 oxygen tokens $O_1,...,O_{14}$; this is visualized in \Cref{fig:arch} within the box labeled ``Tokenizer.''
We translate the input positions $r$ so it has mean $\mu(r)=0$ and subsequently rotate it so it has a diagonal covariance matrix $\Sigma(rR)=\text{diag}(\sigma_{11},...,\sigma_{NN})$ (not included in \Cref{fig:arch} for simplicity). 
We then create a list of atomic orbital token indices $\widehat{z}\in\ZZ^{m}$ and $\widehat{r}\in\RR^{m\times 3}$ where $m$ is the total number of atomic orbitals used to represent the input molecule (e.g. 24 for $H_2O$ used as an example in \Cref{fig:arch}). 
The input $x=\text{embedding}[\widehat{z}]+\widehat{r}\cdot W_{proj}\in\RR^{m\times d_{model}}$ is passed to a vanilla Transformer Encoder \cite{transformer}. 
Since we use the final layer's attention matrix 
 $A$ for prediction, we employ a single attention head and set $Q=K$ so the output $A$ is symmetric, $A=A^T$. 
 
We use $H=H_{init}+A$ to compute energy $E(\cdot)$ as illustrated in \Cref{fig:arch} within the box labelled ``Compute Energy with DFT'' in \Cref{fig:arch}). 
We precompute the necessary DFT tensors using the DFT library PySCF \cite{pyscf} on the CPU asynchronously in a PyTorch dataloader.

\paragraph{Using an Initial DFT Guess.} DFT libraries like PySCF speed up convergence by using an initial guess for the density matrix $\rho$. Specifically, PySCF \cite{pyscf} uses the ``minao initialisation" $\rho_0$ \cite{minao1,minao2}. 
We incorporate this initial guess to our Transformer by using $H_{init}=H(\rho_0)$ where $H(\rho_0)$ is the Hamiltonian computed from $\rho_0$ using Roothaan's equation \cite{roothaan}. We found this to improve optimization convergence. 

\paragraph{Quantum Biased Attention.} Inspired by Transformer-M \cite{transformerm}, we bias the (non-output) pre-softmax attention matrices
 by all pairs of dot products $P_{ij}=\widehat{r}_i^T\widehat{r}_j$ and six cheap-to-compute DFT matrices $H_{core},L^{-1}H_{core}L^{{-1}^T},\rho_0,(J-K/2),V_{xc},H(\rho_0)$. 
 \footnote{The matrix $L^{-1}H_{core}L^{{-1}^T}$ appear in DFT when translating the generalized eigenproblem $\textsc{scipy.linalg.eigh}(H_{core}, O)$ into an eigenproblem $\textsc{np.linalg.eigh}(L^{-1}H_{core}L^{{-1}^T})$ by using $L=\textsc{np.linalg.cholesky}(O)$. }
Our Transformer has $n_{heads}=16$ and we apply each DFT matrix to 2 different attention heads, so the first $14$ heads are biased and the remaining $2$ are ``free.''

\paragraph{Density Mixing.} We found using the DFT technique ``density mixing'' to compute $E(\cdot)$ improved performance. 
\Cref{fig:arch} shows the content of a DFT iteration inside the ``Compute Energy'' box. Density mixing changes this to: 
$$\rho=\alpha \rho+(1-\alpha)2CC^T\quad \alpha\in[0,1).$$
We like to think of density mixing as a residual connection and choose $\alpha=0.5$. Density mixing may break the idempotency $\rho^2=\rho$ needed to compute $E(\cdot)$. We fix this by running yet another DFT iteration. 
In summary, our method use one DFT to initialize $H(\rho_{0})$ followed by another three DFT iterations to stably compute $E(\cdot)$. One might thus view QPT as a hybrid ML/DFT method, with a NN sandwiched between an initial DFT iteration followed by three refining DFT iterations, all trained by backpropagating through DFT. 

\begin{table*}[h!]
\centering
\begin{tabular}{llcccccc}
\hline
Dataset & Method & XC & $\Delta E$ [meV] $\downarrow$ & $\Delta H$ [$10^{-6}E_h$] $\downarrow$ & $\Delta \varepsilon$ [$10^{-6}E_h$] $\downarrow$ & Parameters [M] $\uparrow$ \\ \hline
Water  & SchNOrb & PBE & 1.435 & 165.37 & 279.3  & - \\
       & PhiSNet & PBE & - & 7.87 & 29.81  & 3.839M  \\
       & QHNet & PBE & - & 10.49 & 31.62  & 2.419M \\
       & \textbf{QPT (ours)} & B3LYP & 0.052 & 38.06 & 202.03 & 302M \\
\hline 
Ethanol & SchNOrb & PBE & 0.361 & 187.4 & 334.4  & - \\
        & PhiSNet & PBE & - & 19.60 & 101.10 & 8.673M \\
        & QHNet & PBE & - & 20.45 & 81.29 & 3.917M \\
        & \textbf{QPT (ours)} & B3LYP & 0.073 & 18.49 & 148.90  & 302M\\
\hline 
Malon-    & SchNOrb & PBE & 0.353 & 191.1 & 400.6  & - \\
dialdehyde& PhiSNet & PBE & - & 21.89 & 104.76 & 8.869M \\
          & QHNet & PBE & - & 22.07 & 86.04 & 3.925M \\
          & \textbf{QPT (ours)} & B3LYP & 2.006 & 81.29 & 697.43  & 302M \\
\hline 
Uracil & SchNOrb & PBE & 0.848 & 227.8 & 1760  & 93M  \\
       & PhiSNet & PBE & - & 18.89 & 146.14& 9.135M \\
       & QHNet & PBE & - & 20.33 & 113.48& 4.049M \\ 
    & \textbf{QPT (ours)} & B3LYP & 0.524  & 39.46 & 409.21  & 302M\\ \hline
\end{tabular}
\caption{
Validation error of different molecules.
\citet{schnorb} used def2-SVP/PBE, we used def-SVP/B3LYP to make $E(\cdot)$ differentiable. 
We measure the mean absolute error (MAE) between DFT and NN for energy $E$, Hamiltonian $H$ and molecular orbital energies $\epsilon$. Notably, QPT obtain comparable performance than prior work, while bypassing the huge cost to precompute datasets.  }
\label{table:accuracy}
\end{table*}

\section{Experiments}
The goal of this section is to demonstrate that QPT can achieve comparable performance to prior work while bypassing the cost of DFT dataset creation. 
We compare QPT to \citet{schnorb}, \citet{phisnet} and \citet{qhnet}, which train neural networks to predict the Hamiltonian matrix using supervised learning, i.e., each example consists of $(X_i, H^*(X_i))$ as in \Cref{eqn:classicml}. They train a separate neural network on four different molecules $X=\{\text{water}, \text{ethanol},\text{malondialdehyde}, \text{uracil}\}$ with training sets of sizes $\{500, 25k, 25k, 25k\}$. 
We also train one QPT for each molecule, but use the ``implicit DFT loss'' \Cref{equ:implicitdft} computing $E(\cdot)$ on the fly (i.e., we never use the precomputed labels $H^*(X_i)$). 
In \Cref{table:time}, we compare the dataset creation time next to neural network training time, as reported in prior work \cite{schnorb,phisnet,qhnet}. 

We trained a 302M QPT to minimize \Cref{equ:implicitdft} using the same $X_i$ as \citet{schnorb} without using the labels $H(X_i)$. 
This requires a differentiable implementation of $E(\cdot)$. 
Our implementation is based on \citet{qm1b} and thus use def2-SVP/B3LYP instead of def2-SVP/PBE as done in \citet{schnorb}. \footnote{B3LYP is more computationally demanding. It computes $K$ from \Cref{fig:arch} which PBE ``skips.''}
After training QPT no more than 52h on a single hardware accelerator, it achieved comparable accuracy to prior work (see \Cref{table:accuracy}). 
For example: on Uracil, QPT took 31h to train compared to 626+160h=786h for SchNOrb. 
Notably, the Mean Absolute Error (MAE) of QPT relative to PySCF was 0.524meV for energy $E$, $39.46\mu$Ha for Hamiltonian $H$ and 409.21$\mu$Ha for molecular orbital energies $\epsilon$. 
This is comparable to prior works, i.e., slightly better (lower) than SchNorb and slightly worse (larger) than QHNet and PhiSNet. 
Notably, QPT is never given any labels during training, its predictions are infered by optimizing $E(\cdot)$ (just like DFT). 
QPT performed worst on malondialdehyde, even though we trained it for longer and added another density-mixed~DFT~iteration.

\paragraph{Training Details.} On all tasks, we trained a 302M Transformer following ``GPT medium'' as closely as possible, i.e., $d_{\text{model}}=1024,n_{\text{heads}}=16,n_{\text{layers}}=24$. We used the Adam optimizer \cite{adam} with $\beta_1=0.9,\beta_2=0.95$, 0.1 weight decay and the LR schedule from \href{https://github.com/karpathy/nanoGPT}{minGPT} with 100 warmup steps, $lr=10^{-4}/16/\sqrt{5}$ decaying to $lr=10^{-6}/16$ at step 200k.
We use a batch size of 1. 
The DFT tensors needed to evaluate $E(\cdot)$ are prepared on the CPU using PySCF run in a PyTorch dataloader. 
Similar to how DFT has 50 iterations, we keep a window of the last 6-8 tensors to avoid throttling the dataloader.

\paragraph{Adjusting DFT Tolerance.}
The energy convergence threshold of PySCF is by default $2.72\cdot 10^{-9}$eV compared to the $10^{-4}$eV error of SchNOrb. 
We suspect it is sufficient that the accuracy of $E(\cdot)$ is ``only'' $10\times$ better than the desired accuracy of our NN, e.g., if we want to train QPT on Uracil to $0.3$ meV accuracy, it is okay that our DFT is $0.03$ meV inaccurate.
Prior to training, we ran PySCF to determine sufficiently accurate DFT options which minimize computational resources.
For example: For Uracil we could reduce the exchange correlation grid level from 3 to 2, which reduced the DFT tensor ``gridAO'' from 
(4, 152320, 132) to
(4, 111896, 132) (roughly 25\% faster). 
We also introduce a threshold on the Electron Repulsion Integrals (ERIs) discarding anything below $10^{-7}$, which reduced the number of distinct ERIs from 34.7M to 25.7M (roughly 25\% faster). 
Finally, while $E(\cdot)$ was evaluated in float64, we evaluated $\mlp$ in float32. Similar to how DFT implmentations like TeraChem \cite{terachem} employ variable grid sizes between DFT iterations, we conjecture NNs trained with the implicit DFT loss can utilize adaptive grid sizes. This opens an interesting avenue for future research, exploring how the plethorea of optimization techniques from DFT might translate to hybrid ML/DFT methods (DIIS, density mixing, minao initialization, ERI symmetries and~sparsity,~...).

\begin{figure}[!h]
    \centering
    \includegraphics[width=0.50\textwidth]{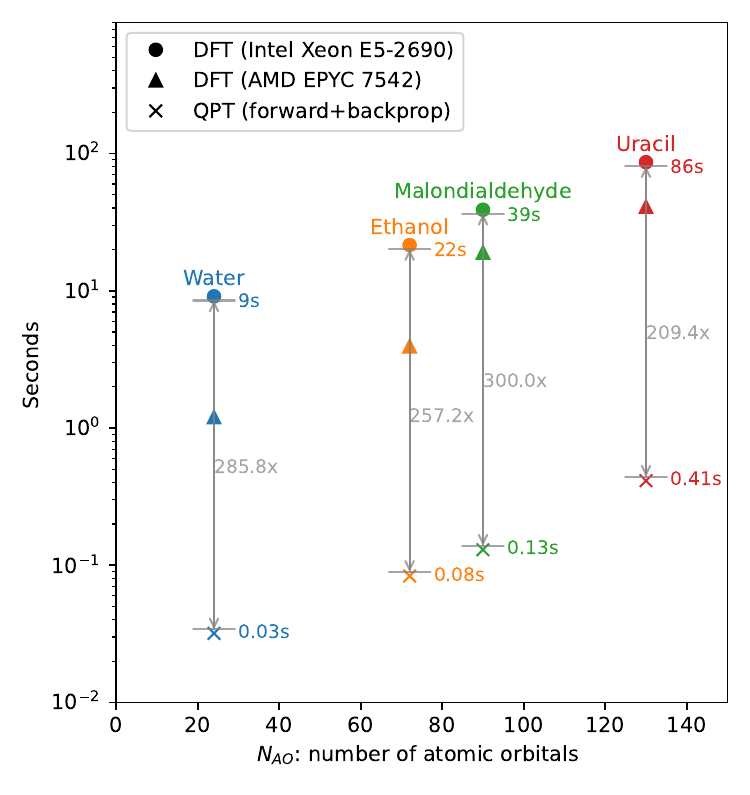}
    \caption{Time to label a training example using DFT as reported by \citet{schnorb}, \citet{phisnet} and \citet{qhnet}, compared to a forward/backward pass of QPT using energy $E(\cdot)$ as a loss function. 
    Bypassing the expensive labeling gives us the ability to evaluate $E(\cdot)$ on a new $X_i$ each iteration.  Triangle is time of DFT using our newer CPUs. 
    }
    \label{fig:traintime}
\end{figure}

\paragraph{Comparing Training Speed to Data Generation} In \Cref{fig:traintime}, we compare the time to label a training example as reported in prior work\footnote{
The numbers in \citet{phisnet} used an old 32nm CPU Intel Xeon E5-2690. We also report numbers on our 7nm AMD EPYC 7542 with PySCF in def2-SVP/B3LYP with default options.}, compared to the time to compute and backpropagate through QPT with the ``implicit DFT loss.''
The time to perform a DFT computation scales with the cube of the number of atomic orbitals $O(N_{AO}^3)$. 
Using DFT to compute the energy loss $E(\cdot)$ amounts to a few DFT iterations and thus also scale $O(N_{AO}^3)$ (backpropagation takes only a constant amount of time longer). 
In practice, we found the time to evaluate and backpropagate through $E(\cdot)$ is around 
 $200\times$ faster than creating a training example by running DFT as reported by prior work. 
This reduction seem to arise from several factors: fewer DFT iterations, hardware acceleration, lower DFT accuracy tuned to $\mlp$, efficient JAX sparse matmul implementation, pre-computing DFT tensors asynchronously on the CPU in the PyTorch dataloader.

\paragraph{Adjusting DFT Iterations. }Early in training, we used a single DFT iteration to compute $E(\cdot)$ as in \Cref{fig:arch}. 
We add an additional iteration with density mixing after 150k steps, before the learning rate scheduler finishes. We add a final DFT iteration for validation to ensure idempotency of the density matrix $\rho^2=\rho$. 
The training speed numbers reported in \Cref{fig:traintime} assume the worst case, i.e., four DFT iterations. To our surpise, using four DFT iterations throughout training performed worse than incrementally adding the DFT iterations. 

\begin{table*}[h!]
    \centering
    \begin{tabular}{|l|cccc|ccc|c|}
\hline
& \multicolumn{4}{c|}{\textbf{DFT}} & \multicolumn{3}{c|}{\textbf{NN}} & \textbf{DFT/NN hybrid} \\
\textbf{Molecule} & \textbf{SchNOrb} & \textbf{PhiSNet} & \textbf{QHNet} & \textbf{Us} & \textbf{SchNOrb} & \textbf{PhiSNet} & \textbf{QHNet} & \textbf{QPT} \\
\hline
Water           & 9.10s  & -    & 11.38s & 1.2s  & 0.033s & -    & 0.0109s & 0.011s (0.050s) \\
Ethanol         & 22.2s & 21.6s & 25.1s  & 3.9s & 0.033s & 0.027s & 0.0711s & 0.040s (0.365s)\\
Malondialdehyde & 43.6s & 38.9s & 40.6s  & 19s   & 0.034s & 0.029s & 0.0792s & 0.065s (0.853s) \\
Uracil          & 88.4s & 86.6s & -     & 41s   & 0.038s & 0.050s  & -      & 0.188s (1.729s) \\
\hline
\end{tabular}
    \caption{Inference numbers. We report numbers ``$x$s ($y$s)'' where $x$ is the time for an inference of QPT and its DFT iterations, while $ys$ is the total time which includes preparing DFT tensors in our dataloader.  The time to prepare the DFT tensors take $10\times$ longer than running QPT and the DFT iterations. We can hide this during training by keeping a window of the last 6-8 molecules, however, this is not possible during inference. }
    \label{tab:inference}
\end{table*}

\paragraph{Inference Speed.} 
We compare the inference time of QPT against DFT and previous NNs in \Cref{tab:inference}. 
We report both the inference time due to QPT (including its DFT iterations) and total time including the dataloader preparing DFT tensors. 
The dataloader increase time consumption almost $10\times$, e.g., from 0.188s to 1.729s for Uracil. 
During training, we mitigated this issue by keeping the last 6 to 8 molecules in RAM and repeating training on them 6-8 times. This is not possible during inference (each molecule is seen only once).

\begin{figure*}[t]
    \centering
    \includegraphics[width=\textwidth]{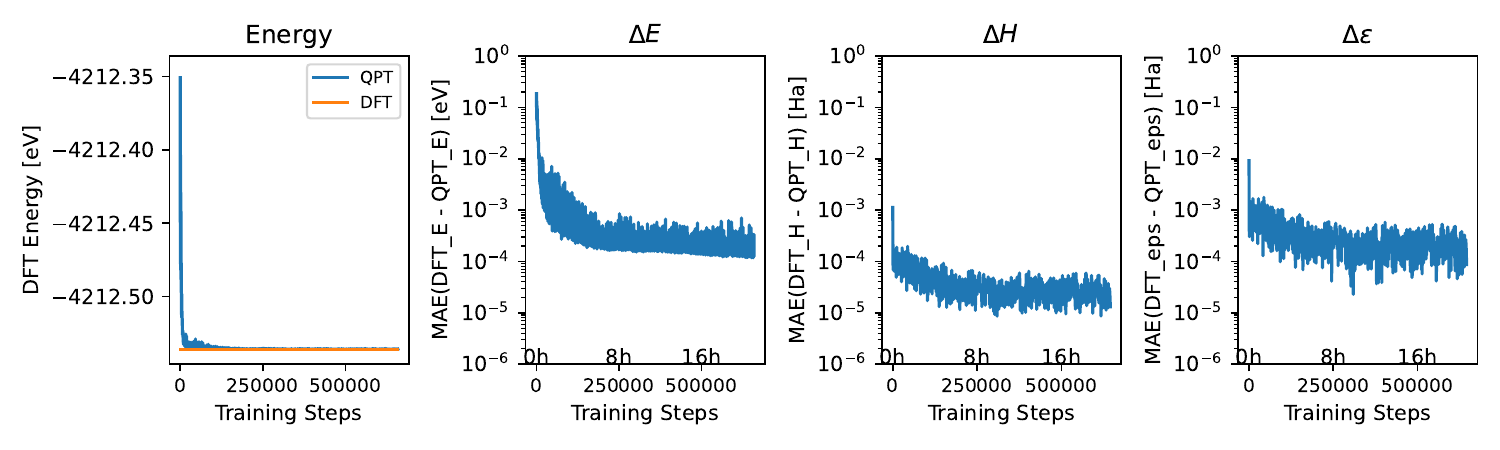}
    \vspace{-2mm}
    \caption{On the left, we visualize $QPT_E$ during training on a single validation example, for which we also computed $DFT_E$. Next, we visualize their difference $\Delta E$. We finally present a similar plot for the resulting Hamiltonian $H$ and its molecular orbital energies $\epsilon$. }
    \label{fig:loss_curve}
\end{figure*}

\paragraph{Loss curves.} \Cref{fig:loss_curve} plots the error of a single ethanol validation molecule as training QPT progress. 
From left: The first plot visualises the energy computed by DFT $DFT_E$ as a horizontal line, and how the energy predicted by QPT $QPT_E$ gets closer as training progress. 
As the lines get closer it becomes hard to see their difference, so the next plot instead visualises their absolute difference 
$|DFT_E - QPT_E|$. 
We similarly plot the mean absolute error (MAE) of the validation molecule for the Hamiltonian matrix, and the molecular orbital energies computed as the eigenvalues from the generalized eigenproblem $\textsc{scipy.linalg.eigh}(H,O)$.

\begin{figure}[h!]
    \centering
    \includegraphics[width=0.3\textwidth]{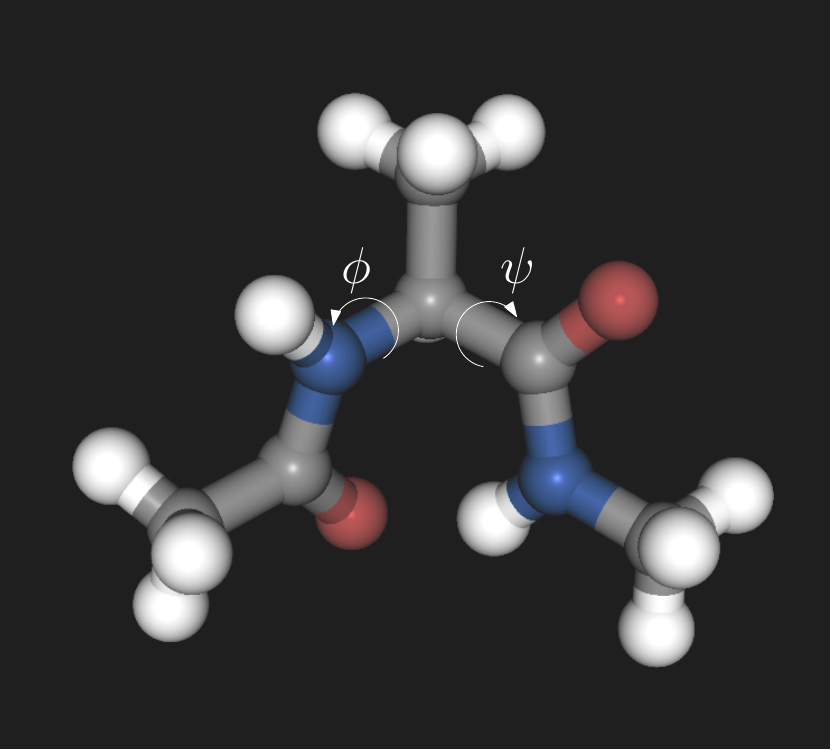}
    \includegraphics[width=0.48\textwidth]{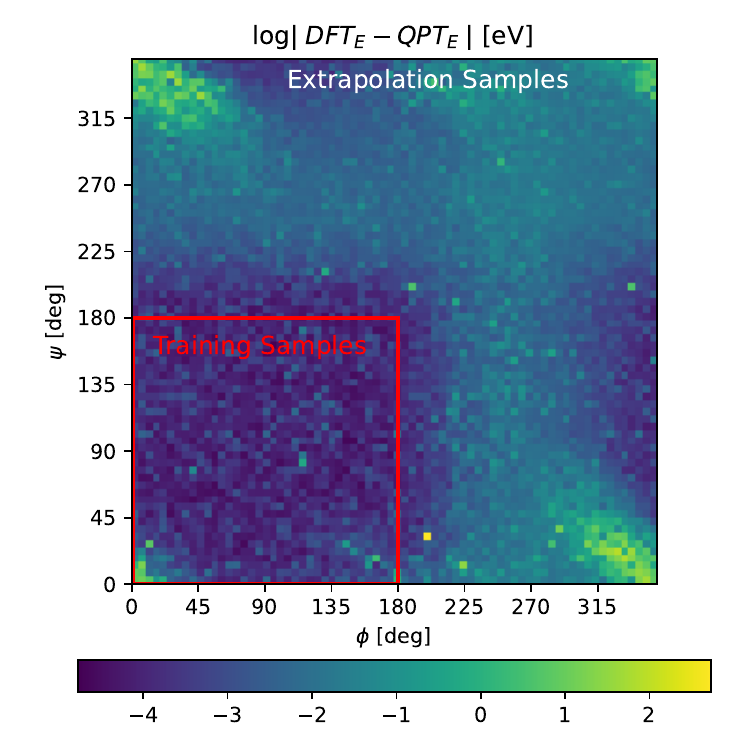}
    \caption{Comparison of the models performance on conformations inside the training distribution $(\phi,\psi)\in [0,180]^2$ relative to the performance outside the training distribution $(\phi,\psi)\in[0,360]^2\backslash[0,180]^2$. 
    The model exhibits minor extrapolation to angles $[180,200]$ not seen during training.  
    A typical energy value is around $13000$eV, so an error $\log(|DFT_E-QPT_E|)=-4$ could mean that the neural network predicted $13000.0001$ instead of $13000.0000$. Chemical accuracy is $0.040$eV). }
    \label{fig:heatmap}
\end{figure}

\paragraph{Preliminary Scaling to Peptides.}
As a preliminary proof-of-concept for proteins/peptides, we trained a QPT on different conformations of alinine dipeptide, represented by two angles $(\phi,\psi)$ (see \Cref{fig:heatmap}). 
During training, we create random conformations by picking two angles $(\phi,\psi)$ uniformly at random from $[0,180]^2$. 
We then measure validation performance within $[0,180]^2$ and extrapolation performance on angles not seen during training (i.e. $[0,360]^2\backslash [0,180]^2$). 
Since each conformation is represented by a 2d point $(\phi,\psi)$, we can visualize the error $\log |DFT_E(\phi,\psi)-QPT_E(\phi,\psi)|$ in a 2d coordinate system (see \Cref{fig:heatmap}). 
The model exhibits minor extrapolation; the error within $[0, 200]^2\backslash[0,180]^2$ is similar to the error within the training distribution $[0,180]^2$. 
To speed up experimentation we use the less accurate DFT options STO3G/B3LYP as done in \cite{qm1b}. This QPT experiment did not include the quantum biased attention and the minao initialization; we expect re-training with these techniques would improve performance.

\section{Limitations and Future Work}

\paragraph{DFT options.} Our main experiments demonstrate QPT with the implicit DFT loss can achieve error relative to def2-SVP/B3LYP that, in absolute value, is similar to the error QHNet/SchNOrb/PhiSNet achieve relative to def2-SVP/PBE. 
While the def2-SVP/B3LYP options have been used by related work \cite{qh9}, the differences between the PBE and B3LYP functionals are significant. \footnote{An excellent overview covering the taxonomy of different functionals can be found in \citep{whichxc}}
The B3LYP functional is more computationally intense since it requires computing the exact Hartree-Fock exchange (the $K$ matrix in \Cref{fig:arch}). 
We estimate using PBE would halve our training time, since roughly 45\% of the training time was spent computing $K$.

\paragraph{Inference Speed. } Our inference speed is currently 10x to 50x worse than prior work. 
The main issue lies in the precomputation of DFT tensors. It turns out the most of this time is spent preparing indices for the sparse einsum, not computing the actual DFT tensors (integrals, XC grids, etc). 
We estimate fixing the issue will improve inference time 5x. 

\paragraph{Generating New Data.} The implicit DFT loss allows us to evaluate $E(\cdot)$ on a new $X_i$ every step of training. 
Never seeing the same training example, as is sometimes the case for Large Language Models, practically solves overfitting. 
In our \Cref{table:accuracy} experiments, we did not use this ability. 
The reason is that the benchmark data has a particular distribution of $X_i$'s which we were unable to reproduce. 
When scaling to large pre-training runs with billions of parameters and hundreds of hardware accelerators, we expect controlling the generation of $X_i$ will be of immense importance. 

\paragraph{Inductive Bias.} This work demonstrate a simple QPT architecture without inductive biases can obtain comparable performance to prior work. It remains to be shown whether inductive biases can improve the performance of models trained with the ``implicit DFT loss.'' We find it particularly interesting whether inductive biases improve or worsen potential scaling laws.

\paragraph{Scaling to Protein-ligands.} 
We tested our current implementation for molecules no larger than $N_{AO}=132$. 
This is very small compared to protein-ligand interactions which contain around 4000 atoms, i.e. $N_{AO}<20k$ in STO3G or $N_{ao}<52k$ in def2-SVP. 
Scaling to this size presents a formidable challenge. 
In our alanine dipeptide we experimenting with a batching technique. 
While we found no computational benefits at the scale of the dipeptide, we conjecture its benefit will improve at protein-scale. 
The idea is to create a batch of similar molecules: fix the protein ($N_{AO}=19900$) and create 200 training examples by moving only the ligand ($N_{AO}=100$). 
To prepare the DFT tensors for the batch with 200 examples, we only need to compute the interactions within the proteins atomic orbitals once; the cost of preparing such 200 training examples is therefore almost the same as preparing 1 protein. \footnote{One can augment moving the ligand with changing the atom types $C\leftrightarrow O$ and $N\leftrightarrow F$.}

\section{Related Work}

\subsection{Supervised Neural Approximations to DFT}
A large body of prior work train neural networks to approximate outputs of DFT \cite{phisnet,qhnet,qh9,schnorb,schnet,mace}. 
Some train their NN to predict the Hamiltonian matrix \cite{phisnet,qhnet,schnorb}, others train to predict properties like homo-lumo gap \cite{schnet,Masters2022GPSAO} while yet others learn to approximate forces \cite{mace,nequib}. 
Notably, while it may be faster to directly predict quantities like forces, both homo-lumo gap and forces can be derived from the Hamiltonian. This line of work differs to ours, in that they all train neural networks with supervised learning using datasets with precomputed labels. We instead propose to use $E(\cdot)$ as a loss function. 

The need for labeled datasets has sparked a large amount of previous work creating public datasets: QM9 \cite{qm9}, ANI1x \cite{ani1x}, Hamiltonian MD17 \cite{schnorb}, PCQ \cite{pcq}, OpenCatalyst \cite{opencatalyst}, QM1B \cite{qm1b}, nablaDFT \cite{nabladft} and many more.
The time to compute these datasets can be very huge, for example, it took a reported 2 years to generate the PCQ dataset \cite{pcq}. Furthermore, for datasets that store DFT matrices, data storage also becomes burdensome, e.g., nablaDFT contains 1M molecules and consumes 7TB of storage. This storage requirement is circumvented when training with the implicit DFT loss (note that a 1M protein dataset with $N_{AO}=50k$ would take up 20PB in float64). 

Finally, the majority of prior work focus on either dataset creation or neural network training. 
Our work differs, in that we join the optimization involved in data creation and neural network training. 
This opens interesting new avenues of merging prior DFT optimization techniques, specifically, tailoring the DFT optimization to neural network pre-training. 
We preliminarily explored adjusting the DFT accuracy by changing the XC grid size, the electron repulsion threshold and the number of DFT iterations used to compute $E(\cdot)$.

\subsection{Differentiable DFT}
The usual DFT algorithm relies on evaluating several derivatives. Most DFT libraries implement derivatives manually instead of using automatic differentiation. 
This is in contrast with the approach adopted in machine learning where frameworks such as PyTorch \cite{pytorch}, Jax \cite{jax}, and Tensorflow \cite{tf} provide powerful automatic differentiation capabilities. 
This has motivated a line of work which reimplement DFT within modern machine learning libraries, allowing automatic differentiation, and, in certain cases, even hardware acceleration \cite{Li2020KohnShamEA,pyscfad,dqc,qm1b,d4ft}. 

Notably, \citet{Li2020KohnShamEA}, \citet{dqc} and \citet{graddft} use their differentiable DFT to improve the accuracy of DFT by training a neural exchange correlation functional by backpropagating through DFT.  Our work differs, in that we do not train exchange correlation functionals to improve the accuracy of DFT. Instead, we use a differentiable DFT implementation of $E(\cdot)$ to interleave the optimization of neural network and DFT $\min_\theta \sum_{i=1}^D E(X_i, \mlp(X_i,\theta))$.

\citet{pyscfad} introduced a plugin to PySCF which provides automatic differentation by replacing NumPy \cite{numpy} with the compatible \texttt{jax.numpy} \cite{jax} modules, however, they remain reliant on libcint and libxc so do not support hardware acceleration. \citet{qm1b} introduced a small subset of PySCF with hardware acceleration, and, sufficient automatic differentiation for our use-case, i.e., we can compute $\nabla_H E(H)$. This led us to base our implementation of \citet{qm1b}. 

\citet{d4ft} revisited DFT from a modern deep learning perspective. In particular, they demonstrate that the DFT optimization problem $\min_{M_i} E(X_i, M_i)$ can be minimized using Gradient Descent with the Adam optimizer. 
One might view our work as a continuation of their work. Instead of optimizing $\min_{H_i} E(X_i, H_i)$ with respect to one matrix $H_i$ for each molecule $X_i$, we optimize neural network parameters $\phi$ for many molecules $\min_\theta \sum_{i=1}^D E(X_i, \mlp(X_i, \theta))$. We then compare the resulting neural network $\mlp $ against those trained with supervised learning \cite{schnorb,phisnet,qhnet}.

\section{Conclusion}
Prior work spend more time labeling datasets with DFT than subsequent NN training. 
For example, it took years to create the PCQ datast \cite{pcq} on which NNs are usually trained within a week. 
We demonstrated training QPT with ``DFTs loss function'' yields comparable accuracy to prior work (\Cref{table:accuracy}), while mitigating the huge dataset creation cost (\Cref{table:time}).  For example, on the Uracil dataset, QPT achieved comparable accuracy to SchNOrb within 31h, whereas SchNOrb was trained for 160h on a dataset which took 626h to create (i.e. a total of 786h).

\clearpage 


\bibliography{main}
\bibliographystyle{style}

\end{document}